\documentclass[10pt, journal]{IEEEtran}

\usepackage{graphicx}

\usepackage[hidelinks]{hyperref}
\usepackage[free-standing-units,per-mode=symbol,mode=text]{siunitx}
\usepackage{amssymb,amsfonts}
\usepackage{amsmath}
\usepackage[export]{adjustbox}
\usepackage{float}
\usepackage{tabularx}
\usepackage{textgreek}
\usepackage{url}
\usepackage[noadjust]{cite}
\usepackage{orcidlink}
\usepackage[acronym]{glossaries}
\usepackage[subtle]{savetrees}
\usepackage{diagbox}
\usepackage{makecell}

\newcommand{\circled}[1]{\raisebox{.5pt}{\textcircled{\raisebox{-.9pt} {#1}}}}

\newcommand{\tinyml}{tinyML}
\newcommand{\frontend}{\textit{Frontend}}
\newcommand{\middleware}{\textit{Midend}}
\newcommand{\backend}{\textit{Backend}}
\newcommand{\platform}{\textit{Deployment Platform}}
\newcommand{\variablebuffer}{\textit{Variable Buffer}}
\newcommand{\transientbuffer}{\textit{Transient Buffer}}
\newcommand{\constantbuffer}{\textit{Constant Buffer}}
\newcommand{\memorylevelannotation}{\textit{Memory Level Annotation}}
\newcommand{\typeinference}{\textit{Type Inference \& Kernel Selection}}
\newcommand{\tiling}{\textit{Tiling \& Memory Scheduling}}
\newcommand{\neureka}{\textit{N-Eureka}}

\newcommand{\scenarioOne}{\textit{Single Core Deployment}}
\newcommand{\scenarioTwo}{\textit{Octa-Core Deployment}}
\newcommand{\scenarioThree}{\textit{\gls{npu} without \gls{nms} Deployment}}
\newcommand{\scenarioFour}{\textit{\gls{npu} with \gls{nms} Deployment}}

\newcommand{\autoregressivemode}{autoregressive inference mode}
\newcommand{\parallelmode}{parallel inference mode}
\newcommand{\prompt}{prompt}

\definecolor{darkred}{HTML}{A8322C}
\newcommand{\revA}[1]{\textcolor{black}{#1}}

\newacronym[plural=DNNs, firstplural=Deep Neural Networks (DNNs)]{dnn}{DNN}{Deep Neural Network}
\newacronym[plural=SoCs, firstplural=Systems-on-chip (SoCs)]{soc}{SoC}{System-on-chip}
\newacronym[plural=ONNXs, firstplural=Open Neural Network Exchanges (ONNXs)]{onnx}{ONNX}{Open Neural Network Exchange}
\newacronym[plural=TCFs, firstplural=Tile Constraint Flows (TCFs)]{tcf}{TCF}{Tile Constraint Flow}
\newacronym[plural=TCs, firstplural=Tile Constraints (TCs)]{tc}{TC}{Tile Constraint}
\newacronym[plural=GEMMs, firstplural=General Matrix Multiplications (GEMMs)]{gemm}{GEMM}{General Matrix Multiplication}
\newacronym{cp}{CP}{Constraint Program}
\newacronym[plural=QNNs, firstplural=Quantized Neural Networks (QNNs)]{qnn}{QNN}{Quantized Neural Network}
\newacronym[plural=FMs, firstplural=Foundation Models (FMs)]{fm}{FM}{Foundation Model}
\newacronym{xr}{XR}{Extended Reality}
\newacronym[plural=RNNs, firstplural=Recurrent Neural Networks (RNNs)]{rnn}{RNN}{Recurrent Neural Network}
\newacronym[plural=CNNs, firstplural=Convolutional Neural Networks (CNNs)]{cnn}{CNN}{Convolutional Neural Network}
\newacronym{nlp}{NLP}{Natural Language Processing}
\newacronym{hpc}{HPC}{High-Performance Computing}
\newacronym{eeg}{EEG}{Electroencephalography}
\newacronym[plural=LLMs, firstplural=Large Language Models (LLMs)]{llm}{LLM}{Large Language Model}
\newacronym[plural=SLMs, firstplural=Small Language Models (SLMs)]{slm}{SLM}{Small Language Model}
\newacronym[plural=GPGPUs, firstplural=General-Purpose Graphics Processing Units (GPGPUs)]{gpgpu}{GPGPU}{General-Purpose Graphics Processing Unit}
\newacronym[plural=TPUs, firstplural=Tensor Processing Units (TPUs)]{tpu}{TPU}{Tensor Processing Unit}
\newacronym[plural=DMAs, firstplural=Direct Memory Access (DMAs)]{dma}{DMA}{Direct Memory Access}
\newacronym{ai}{AI}{Artificial Intelligence}
\newacronym[plural=MLs, firstplural=Machine Learnings (MLs)]{ml}{ML}{Machine Learning}
\newacronym[plural=MMUs, firstplural=Memory-Management Units (MMUs)]{mmu}{MMU}{Memory-Management Unit}
\newacronym[plural=ISAs, firstplural=Instruction Set Architectures (ISAs)]{isa}{ISA}{Instruction Set Architecture}
\newacronym{simd}{SIMD}{Single Instruction Multiple Data}
\newacronym{spmd}{SPMD}{Single Program Multiple Data}
\newacronym{dsp}{DSP}{Digital Signal Processing}
\newacronym[plural=MCUs, firstplural=Microcontrollers (MCUs)]{mcu}{MCU}{Microcontroller}
\newacronym{tcdm}{TCDM}{Tightly-coupled Data Memory}
\newacronym{nms}{NMS}{Neural Memory Subsystem}
\newacronym{npu}{NPU}{Neural Processing Unit}
\newacronym{mram}{MRAM}{Magnetoresistive Random Access Memory}
\newacronym{sram}{SRAM}{Static Random Access Memory}
\newacronym{axi}{AXI}{Advanced eXtensible Interface Bus}
\newacronym{ptq}{PTQ}{Post-Training Quantization}
\newacronym{qat}{QAT}{Quantization-Aware Training}
\newacronym{efm}{EFM}{Embodied Foundation Model}
\newacronym{ast}{AST}{Abstract Syntax Tree}
\newacronym{tqt}{TQT}{Trained Quantization Thresholds}
\newacronym{api}{API}{Application Programming Interface}
\newacronym{cim}{CIM}{Compute-In Memory}
\newacronym{mps}{MPS}{Metal Performance Shaders}
\newacronym[plural=MPUs, firstplural=Microprocessors (MPUs)]{mpu}{MPU}{Microprocessor}
\newacronym{dram}{DRAM}{Dynamic Random Access Memory}
\newacronym{ann}{ANN}{Artifical Neural Network}

\begin{document}
\bstctlcite{IEEEexample:BSTcontrol}

\title{Deeploy: Enabling Energy-Efficient Deployment of Small Language Models On Heterogeneous Microcontrollers}

%\author{Author hidden for double-blind review}

\author{
        Moritz Scherer~\orcidlink{0000-0002-2762-2307},~\IEEEmembership{Graduate Student Member,~IEEE},
        Luka Macan~\orcidlink{0009-0007-6130-8841},~\IEEEmembership{Graduate Student Member,~IEEE},
        Victor J.B. Jung~\orcidlink{0009-0001-7462-3468},~\IEEEmembership{Graduate Student Member,~IEEE},
        Philip Wiese\orcidlink{0009-0001-7214-2150},~\IEEEmembership{Graduate Student Member,~IEEE},
        Luca Bompani\orcidlink{0009-0002-1277-8584},~\IEEEmembership{Graduate Student Member,~IEEE},
        Alessio Burrello~\orcidlink{0000-0002-6215-8220},~\IEEEmembership{Member,~IEEE},
        Francesco Conti~\orcidlink{0000-0002-7924-933X},~\IEEEmembership{Member,~IEEE},
        Luca Benini~\orcidlink{0000-0001-8068-3806},~\IEEEmembership{Fellow,~IEEE}%
\thanks{M. Scherer, V. JB. Jung, P. Wiese, are with the Integrated Systems Laboratory, ETH Z\"urich, 8092 Z\"urich, Switzerland; e-mail \{scheremo,jungvi,wiesep\}@iis.ee.ethz.ch.}%
\thanks{L. Macan, L. Bompani, and F. Conti are with the Department of Electrical, Electronic, and Information Engineering (DEI), University of Bologna, 40126 Bologna, Italy; e-mail \{luka.macan, luca.bompani5, f.conti\}@unibo.it}%
\thanks{L. Benini is with the Integrated Systems Laboratory, ETH Z\"urich, 8092 Z\"urich, Switzerland and the Department of Electrical, Electronic, and Information Engineering (DEI), University of Bologna, 40126 Bologna, Italy; e-mail lbenini@iis.ee.ethz.ch}%
\thanks{A. Burrello is with the Department of Control and Computer Engineering, Politecnico di Torino, 10129, Turin, Italy; e-mail alessio.burrello@unibo.it}%
\thanks{This work was funded in part by the Spoke 1 on Future HPC of the Italian Research Center on High-Performance Computing, Big Data and Quantum Computing (ICSC) funded by MUR Mission 4 - Next Generation EU, in part by the Chips Joint Undertaking (Chips-JU) TRISTAN project under grant agreement No 101095947, the CONVOLVE project under grant agreement No 101070374, and the EU Horizon Europe project NeuroSoC under Grant 101070634. Chips-JU, CONVOLVE, and NeuroSoC receive support from the European Union’s Horizon Europe research and innovation program.
}
}

\maketitle

\begin{abstract}
With the rise of Embodied Foundation Models (EFMs), most notably Small Language Models (SLMs), adapting Transformers for edge applications has become a very active field of research. However, achieving end-to-end deployment of SLMs on microcontroller (MCU)-class chips without high-bandwidth off-chip main memory access is still an open challenge. In this paper, we demonstrate high-efficiency end-to-end SLM deployment on a multicore RISC-V (RV32) MCU augmented with ML instruction extensions and a hardware neural processing unit (NPU). To automate the exploration of the constrained, multi-dimensional memory vs. computation tradeoffs involved in aggressive SLM deployment on heterogeneous (multicore+NPU) resources, we introduce Deeploy, a novel  Deep Neural Network (DNN) compiler,  which generates highly-optimized C code requiring minimal runtime support. We demonstrate that Deeploy generates end-to-end code for executing SLMs, fully exploiting the RV32 cores' instruction extensions and the NPU: We achieve leading-edge energy and throughput of \SI{490}{\micro\joule \per Token}, at \SI{340}{Token \per \second} for an SLM trained on the TinyStories dataset, running for the first time on an MCU-class device without external memory.

\end{abstract}

\glsresetall

\IEEEpeerreviewmaketitle

\section{Introduction}\label{sec:intro}

The latest evolutions in mainstream \gls{ai} have been driven by Transformers, which have taken over from \glspl{rnn} and \glspl{cnn} as the leading edge models for language processing and multi-modal applications~\cite{touvron_llama_2023, jiang_mixtral_2024}. The success of Transformers can be primarily attributed to the emergence of the \gls{fm} paradigm: large Transformer models extensively pre-trained on datasets spanning trillions of tokens and then fine-tuned with a much lower volume of labeled data to solve domain-specific problems.
Following the success of \glspl{fm} in \gls{nlp}~\cite{touvron_llama_2023, geminiteam_gemini_2023}, an increasing number of fields are starting to formulate and adapt \glspl{fm} for high dimensional sensor data that has traditionally been challenging to process, like decoding neural data~\cite{chen_eegformer_2024, wang_brainbert_2022}, or training embodied \gls{ai} agents~\cite{ahn_can_2022, driess_palme_2023}, which may incorporate multi-modal sensor inputs. 

Operating directly on sensory data and in a cyber-physical loop may lead to solving many outstanding challenges in fields such as brain-machine interfaces~\cite{wang_brainbert_2022} and miniaturized robotics~\cite{driess_palme_2023}.
However, to materialize this promise, models of this class need to be \textit{embodied} in physical devices as \glspl{efm}, and they must cope with the strict constraints in terms of compute throughput, power consumption, and footprint typical of edge devices.
Unlike datacenter-scale systems, which collect and aggregate sensor data over sharded resources for high-throughput processing, embodied \gls{ai} systems must process sensor data with extremely low latency and memory capacity under tight power constraints.
This is particularly challenging for the smallest class of \gls{ai}-oriented computers: so-called ``\textit{\tinyml{}}'' devices operating at the extreme edge, based on microcontroller-class devices without complex operating systems or \glspl{mmu}, relying on user-level software to implement low-level hardware management functionalities.
Despite many recent successes with previous-generation \glspl{dnn}, the emergence of the \tinyml{} paradigm for \glspl{efm} faces the dual challenge of reducing \glspl{fm} to a manageable size and enabling their deployment on tiny devices.

A first concrete step in this direction is the recent introduction of \glspl{slm}: \glspl{fm} with tens to a few hundred million, rather than several billion parameters~\cite{eldan_tinystories_2023, zhang_tinyllama_2024}.
While most currently available \glspl{fm} are focused on processing natural language at a proof-of-concept scale, the effort towards embedded multi-modal sensor inputs with small-scale, application-specific \glspl{fm} offers a highly promising path for the development of this novel class of models.
Much like what happened with the initial emergence of Deep Learning~\cite{lecun_deep_2019}, the evolution of advanced \tinyml{} applications based on \glspl{efm} is currently prevented by the lack of suitable targets for deployment of these models and, even more, of deployment frameworks that enable utilizing existing specialized hardware to its full capabilities.

Deploying tiny \glspl{efm} requires overcoming several challenges specific to the \tinyml{} domain.
Large-scale \gls{ai} inference systems typically employ heterogeneous computer architectures composed by a conventional host (e.g., an x86 processor) and a very large throughput-oriented accelerator (e.g., H100~\cite{choquette_nvidia_2023}, TPU~\cite{jouppi_tpu_2023}), which is fully exploited only at large batch sizes.
Conversely, \tinyml{} is used for latency-sensitive applications focusing on real-time inference without batching. As a consequence, \tinyml{} \gls{ai} inference typically employs much more specialized accelerator architectures~\cite{prasad_siracusa_2023, ueyoshi_diana_2022}, leading to more complex mapping and optimization challenges for \gls{dnn} deployment.
Furthermore, \tinyml{}'s strict constraints on energy efficiency and microcontroller-class computer architecture typically require platform-specific optimization, including memory-aware tiling, static memory allocation, and latency-hiding \gls{dma} scheduling, which require advanced compiler support to scale to complex \glspl{dnn} like \glspl{fm}. While several compilers have limited support for user-defined kernels~\cite{lattner_mlir_2021, chen_tvm_2018}, configuring and extending them requires expert knowledge, and their top-down compilation approach often clashes with loosely coupled accelerators. Moreover, mainstream compilers do not address the strict memory constraints in extreme-edge devices.

In this paper, we aim to remove the first barrier towards developing \gls{efm} suited for deployment on \tinyml{} platforms: the lack of deployment frameworks that enable their efficient execution.
We demonstrate, to the best of our knowledge, the first end-to-end tool flow to deploy \glspl{efm} on heterogeneous microcontroller-class systems. Specifically, we demonstrate the end-to-end deployment of a TinyStories-class~\cite{eldan_tinystories_2023} network on \textit{Siracusa}, an advanced microcontroller in TSMC \SI{16}{\nano\meter} technology featuring embedded non-volatile memory (MRAM) and two heterogeneous compute engines, namely, an octa-core RV32 compute cluster with instruction extensions for ML and a multi-mode \gls{cnn} \gls{npu}, \neureka{}~\cite{prasad_siracusa_2023}. We present the tooling and algorithms integrated within our deployment framework, Deeploy.

The contributions of this paper are as follows:
\begin{itemize}
    \item We describe \textit{Deeploy}, a customizable, domain-specific compiler designed for generating bare metal code fitting the memory constraints of extreme edge devices. Deeploy supports all the key computational primitives needed for the execution of Transformer-based \glspl{efm} on heterogeneous extreme edge \glspl{soc} through its bottom-up compilation approach, which allows applying advanced code optimization on expert-optimized kernel templates. We further introduce a novel algorithm for solving the tiling and static memory allocation problems for multi-level software-managed caches and its integration into Deeploy.\footnote{We will open-source all code required to reproduce our experiments under https://github.com/pulp-platform/deeploy}.
    \item We benchmark common Transformer configurations, demonstrating that code generated by Deeploy maximizes engine utilization in heterogeneous, multi-accelerator \glspl{soc}. We achieve data marshaling overheads of just 9\,\% for large workloads with high arithmetic intensity executing on the cluster cores and \gls{npu} collaboratively thanks to efficient data movement acceleration and low-overhead offloading mechanisms.
    \item As a concrete large-scale end-to-end use-case of Deeploy and its adaptability to heterogeneous hardware platforms, we demonstrate for the first time the deployment of a TinyStories-class \gls{slm} on Siracusa, a state-of-the-art heterogeneous \gls{mcu}. While using on-chip memory only, we achieve a throughput of \SI{340}{Token \per \second} at an energy cost of \SI{490}{\micro\joule} for autoregressive inference. We show that using the flexible deployment flow enabled by Deeploy for the same \gls{slm} allows us to implement multi-layer $KV$ caching using on-chip memory only, improving token throughput by 26\,$\times$ compared to inference without caches.

\end{itemize}

The rest of this paper is organized as follows: in Section~\ref{sec:related}, previous work in quantized neural networks, small language models, and neural network deployment for extreme edge devices is introduced and discussed. Section~\ref{sec:deeploy} introduces Deeploy and discusses its deployment flow for Transformers. Section~\ref{sec:tinyLlama} discusses the \gls{slm} architecture used in this work and the approach to mapping it on Siracusa. In Section~\ref{sec:siracusa}, we present the Siracusa \gls{mcu} platform. Section~\ref{sec:results} presents and discusses the end-to-end deployment results, comparing them to the state-of-the-art. Finally, Section~\ref{sec:conclusion} concludes this paper, summarizing the results and contributions.
\section{Related Work}\label{sec:related}

%In the last few years, the emergence of open-source \glspl{fm} has led to a surge of research into applications and optimizations of this class of models. At the same time, \tinyml{} became more attractive for AI-based applications. 
This Section gives an overview of the state-of-the-art on \glspl{efm}, focusing on developments towards improvements in energy efficiency and model size and tools to deploy \glspl{dnn} on extreme edge devices.

\subsection{Small Foundation Models}

Recently, the development of decoder-only \glspl{llm} such as Llama~\cite{touvron_llama_2023}, and Mixtral~\cite{jiang_mixtral_2024}, and their associated \gls{ml} pipelines led to a new model type: the Foundation Model (FM). 
% and GPT~\cite{brown_language_2020} 

\glspl{fm} are pre-trained \glspl{llm}, which can be fine-tuned for downstream tasks at a fraction of the cost of pre-training, making them particularly relevant for domain specialization. 
%These models have shown remarkable success in generating coherent and contextually relevant outputs, much above the capabilities of more traditional architecture like \glspl{rnn}. 
However, \glspl{llm} often contain several billion parameters, requiring \si{\gibi \byte} of storage space, making them incompatible with extreme edge inference.
%, where devices typically offer \SI{10}{\mebi\byte} or less. 

Addressing this gap, the emerging field of \glspl{slm} has gained significant traction in the last year. The aim of \glspl{slm} is to compact \glspl{llm} down to tens to hundreds of \si{\mebi\byte}~\cite{eldan_tinystories_2023, zhang_tinyllama_2024}, mirroring the evolution of compression of \glspl{cnn}~\cite{han_deep_2016} over the past decade.

This paradigm shift towards compact \glspl{fm} is particularly interesting for \tinyml{} applications. Incorporating smaller \glspl{fm}, like \glspl{slm}, into embedded devices may enable a new wave of intelligent, responsive, and autonomous devices built on \glspl{efm}. Such systems could bridge the gap between human-understandable inputs such as text and performing high-level planning and low-level control tasks~\cite{firoozi_foundation_2023} and make such advanced capabilities available at the edge, embodied in robots, appliances, and wearable devices.

In this work, we contribute to the growing field of \gls{slm} and \gls{efm} research and aim to lay the foundation for truly embedded \glspl{slm} by providing a foundational deployment flow that supports a wide range of \glspl{fm}, from autoregressive decoder-only ones to encoder-only ones. 

\subsection{Quantized Transformer Models}\label{sec:quantization}

Neural network quantization has been an active field of research for the past decade, as the promises of reduced parameter storage and higher compute efficiency on reduced-precision operands drive the development of increasingly aggressive quantization methods~\cite{tekin_review_2024, han_deep_2016}.
%~\cite{zhu_survey_2023}

Improvements in energy efficiency are significant when switching from floating-point point computation to integer arithmetic~\cite{gholami_survey_2022, kim_ibert_2021} due to the reduced hardware complexity required to implement the fundamental operations using integer arithmetic. One commonly used approach to quantize \glspl{dnn} is \gls{qat}, where the model is trained to overcome quantization effects that occur when using lower-precision values for weights and activations~\cite{jain_trained_2020}. 
However, \gls{qat} often requires computationally expensive retraining of the model and access to representative datasets, which are not readily available. \gls{ptq} methods can be applied to quantize models without retraining while conserving full-precision accuracy~\cite{ li_brecq_2020}.
Especially in the domain of \glspl{fm}, \gls{ptq} has been successfully applied~\cite{xiao_smoothquant_2023} to reduce the computational cost of quantization.
%dai_vsquant_2021,
In this work, we apply state-of-the-art \gls{ptq} on a publicly available pretrained \gls{slm} to achieve quantized inference without loss of accuracy, a prerequisite for energy-efficient inference on extreme edge devices.

\subsection{Neural Network Deployment for Extreme Edge Devices}\label{sec:deploymentalgorithms}

Building on the trends of model quantization and compression, as well as research into more computationally efficient \glspl{dnn}~\cite{howard_searching_2019}, \gls{dnn} inference on mobile and embedded devices has become a flourishing field of research~\cite{prasad_siracusa_2023, ueyoshi_diana_2022, lin_memoryefficient_2021}. 
%lin_mcunet_2020, 
While model deployment on mobile devices like smartphones follows similar approaches to server-scale deployment, relying on the ample compute- and memory resources, hardware-managed caches, and operating systems to carry out task scheduling available to this class of devices, deeply embedded devices face much more severe constraints in deployment. This is especially true for the new generation of \gls{mcu}-class devices focusing on \gls{ai} applications. In contrast to their predecessors, these \glspl{mcu} feature multi-core compute clusters, \gls{dnn} accelerators, and on-chip memory of up to \SI{10}{\mebi\byte}, split into multiple software-managed memory hierarchy levels~\cite{prasad_siracusa_2023, flamand_gap8_2018, ueyoshi_diana_2022}.

To optimally leverage the compute capabilities of such complex systems, network deployment must simultaneously optimize the execution schedule and tiling of operators and orchestrate overlapping memory transfers using \glspl{dma} to achieve low data marshaling overheads and high compute utilization. 
While modern top-down compilers like MLIR and TVM~\cite{lattner_mlir_2021, chen_tvm_2018} allow integration of most common \glspl{isa} and accelerator APIs, their focus is not on meeting the stringent memory constraints of this class of \tinyml{} devices.
Prior work like Dory~\cite{burrello_dory_2021}, CoSa~\cite{huang_cosa_2021}, and others have addressed these challenges for \glspl{cnn} by focusing on operator tiling to fit the target's memory constraints. However, these approaches assume a single-cluster memory hierarchy, with undivided memory at each level, and a simple lifetime model for network tensors, which are fundamentally stateless across inference rounds. These simplifying assumptions do not hold for complex heterogeneous multi-accelerator hardware and advanced \gls{slm} networks~\cite{maas_telamalloc_2022, moffitt_minimalloc_2024}.

Moving beyond these prior works, we propose a novel constraint programming algorithm that enables co-optimizing tiling and memory allocation, which overcomes the limitations of previous approaches by supporting data flows with complex lifetimes (e.g. $KV$ caching) as required by \glspl{efm}.
\begin{figure*}
\begin{center}
\includegraphics[width=0.975\linewidth]{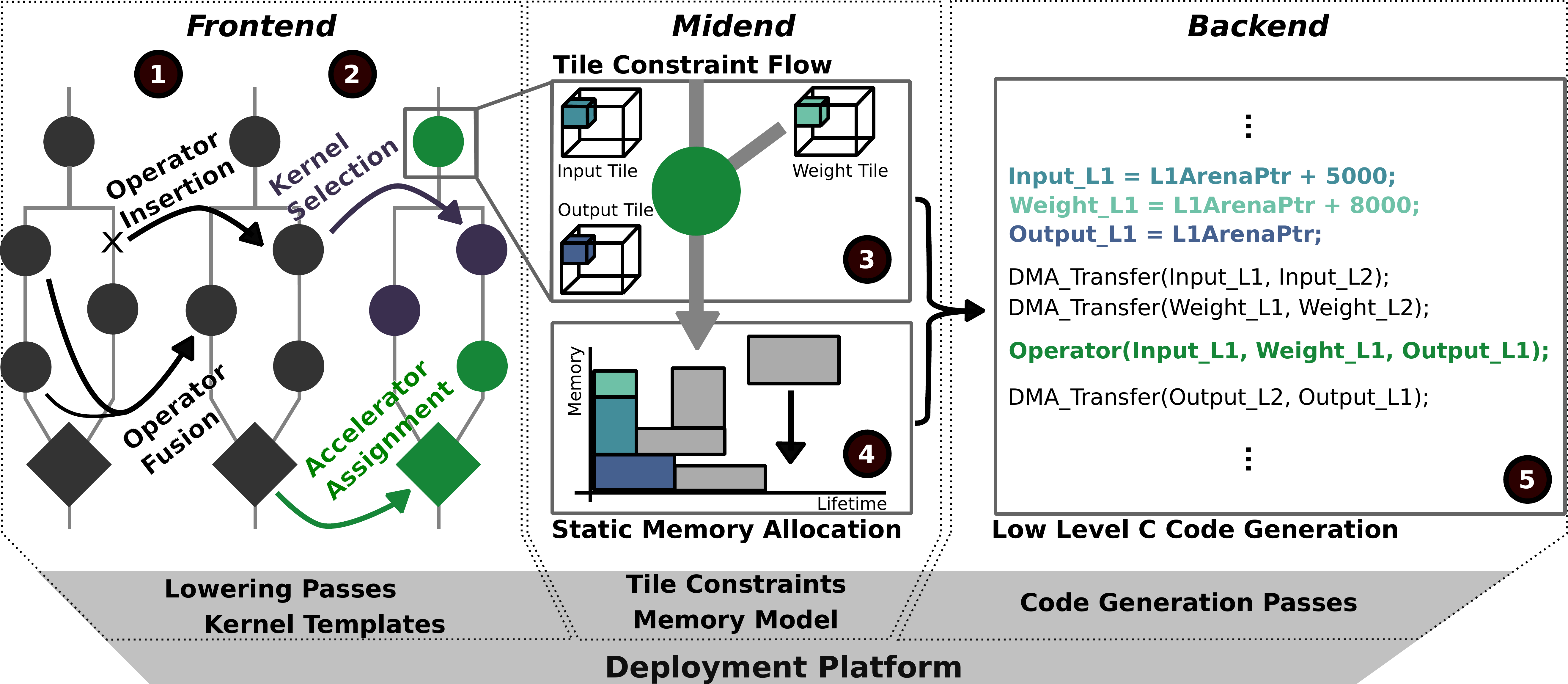}
\caption{Overview of the Deeploy Execution Flow. Steps \circled{1} and steps \circled{2} are part of the \frontend{}. In the first step, the graph is modified by fusing and inserting platform-specific operators, for example, transposition operators, to match data layout requirements. In the second step, datatypes for every tensor are inferred, the accelerator target is chosen, and kernel templates are selected. The first step in the \middleware{}, step \circled{3}, is the Tile Constraint Flow, which computes geometrical constraints for the tile sizes of each tensor, adding them to a \gls{cp}. The resulting tensor size variables are translated into a 2D bin packing problem in step \circled{4}. The solution of the co-constrained tiling and static memory allocation problem is computed by the ORTools CP-SAT solver and finally processed in step \circled{5} in the \backend{}. Step \circled{5} generates platform-specific C Code exploiting \gls{dma} transfers. Each step of the execution flow is highly configurable through the \platform{} object.}
\label{fig:executionflow}
\end{center}
\end{figure*}

\section{Deeploy}\label{sec:deeploy}

In this section, we provide an overview of the Deeploy compilation flow. In contrast to most state-of-the-art compilers for \glspl{dnn}, which lower \gls{dnn} representations top-down into predefined primitives that need to be implemented by each backend~\cite{chen_tvm_2018, lattner_mlir_2021, david_tensorflow_2021}, Deeploy employs a bottom-up compilation approach, where the compiler implements networks by composing user-provided C kernels, extending them with code generation passes to implement tiling and memory allocation.
This bottom-up approach to compilation provides three key advantages: first, it supports reusing hand-optimized kernel libraries commonly available for most \glspl{isa} and accelerators. 
Second, it can be easily extended to support highly customized non-standard compute platforms, including heterogeneous \glspl{soc} featuring multiple accelerators for which a low-level compiler backend may not exist.
Third, it allows easy integration of novel operators found in emerging Transformer architectures without invasive modifications to the deployment flow.

Deeploy is organized in three building blocks; the \frontend{} validates and transforms the graph representation into a representation that suits the platform and assigns kernel templates to each operator. The \middleware{} performs all tiling and static memory allocation computations\revA{, guaranteeing that the computed program schedule may execute without unscheduled runtime memory spills}. Finally, the \backend{} uses the optimized graph representation generated in the \frontend{}, and the generated tiling schedule and memory allocation map generated in the \middleware{} to create executable code through a series of code generation passes. All deployment targets share the same execution flow, and Deeploy uses a configurable platform abstraction, the \platform{}, which allows it to steer operators' mapping, optimization, and lowering according to the platform's configuration. An overview of the Deeploy execution flow is shown in Figure~\ref{fig:executionflow}.

\subsection{Data Structures}\label{sec:datastructures}

Deeploy distinguishes between three types of buffers: \variablebuffer{}s, \transientbuffer{}s, and \constantbuffer{}s. \variablebuffer{}s represent tensors that contain data that is not constant at compile-time, i.e., network inputs, outputs, and intermediate activations. \constantbuffer{}s represent compile-time constant data used in inference, i.e., network weights and other network parameters. Lastly, \transientbuffer{}s represent scratchpad memory locations for kernel execution, e.g., \textit{im2col} buffers for convolution kernels~\cite{garofalo_xpulpnn_2020, lai_cmsisnn_2018}, or reorder buffers for efficient transposition kernels. Typically, the amount of space used in \transientbuffer{}s depends on the operator's parametrization, distinguishing them from \variablebuffer{}s.
In contrast to simpler \gls{dnn} topologies, \glspl{efm} employ data structures that require advanced allocation strategies, such as the $KV$ caches of autoregressive \glspl{slm}, as they have more complex buffer lifetime requirements than intermediate tensors found in \glspl{cnn}.
Addressing these constraints requires a more sophisticated management of the buffers' lifetime and memory allocation than in other deployment tools targeting extreme edge devices~\cite{burrello_dory_2021, huang_cosa_2021}.

The distinction between global and local section buffers is relevant for code generation; global objects are allocated as global C variables, while local objects are only accessible in the inference code. As such, global variables are alive throughout an inference execution, while local variables are allocated and deallocated as the network's execution schedule requires. 

\subsection{Frontend}

Deeploy's \frontend{} is designed around ingesting quantized \gls{onnx} graphs produced by \gls{dnn} and Transformer quantization tools like Quantlib~\cite{spallanzani_quantlab_2022}. Deeploy implements a configurable lowering pass system based on pattern matching of \gls{onnx} graphs to enable efficient and customizable graph-lowering strategies. \revA{Each lowering pass consists of a user-defined replacement function and a \textit{source pattern}, which describes the sub-graph that should be replaced. Using the replacement function, each lowering pass uses the matched sub-graph to generate a \textit{target pattern}, which replaces the \textit{source pattern}.}
Using this system, the first processing step in the \frontend{} is transforming the input graph into a custom, platform-specific \gls{onnx} dialect using lowering passes provided by the \platform{}. \revA{The user further defines operator mappings between custom operators and the engines available in the target platform to control the code generation on the level of individual operators.} Common \tinyml{} kernel libraries like CMSIS-NN and PULP-NN~\cite{lai_cmsisnn_2018, garofalo_xpulpnn_2020} offer kernels for fused linear operators and activations, which can be lowered into by matching pairs of linear operators and quantization operators. Besides operator fusion optimization passes, Deeploy also supports the minimization and insertion of data marshaling operators like transpositions to match the data layout requirements of kernel libraries. An example of such an operator insertion pass is adding transpositions operators to optimize the data layout of the $B$ matrix for \gls{gemm} kernels of type $Y = \alpha AB + \beta C$ for better data access locality.

The second step after transforming the input graph into the platform-specific dialect in the \frontend{} is parsing, during which every operator in the network is analyzed to construct an initial context of buffers used in the network's execution, and \typeinference{} where every buffer in the context is assigned a type. The types used in Deeploy correspond to standard C types (e.g., \textit{int8\_t}, \textit{float32}) or custom data types, depending on the kernels used by the \platform{}. To guarantee a valid type assignment, Deeploy propagates type information top-to-bottom. The user must only provide the input types for every graph's input tensor to achieve this; then,  using this information, Deeploy matches the input types of each operator with one of the kernel signatures provided by the \platform{}.

The final result of the \frontend{} is an assignment of low-level kernel templates to every operator in the lowered platform-specific \gls{onnx}, which satisfies the type constraints imposed by the network's operators.

\begin{figure*}
\begin{center}
\includegraphics[width=0.975\linewidth]{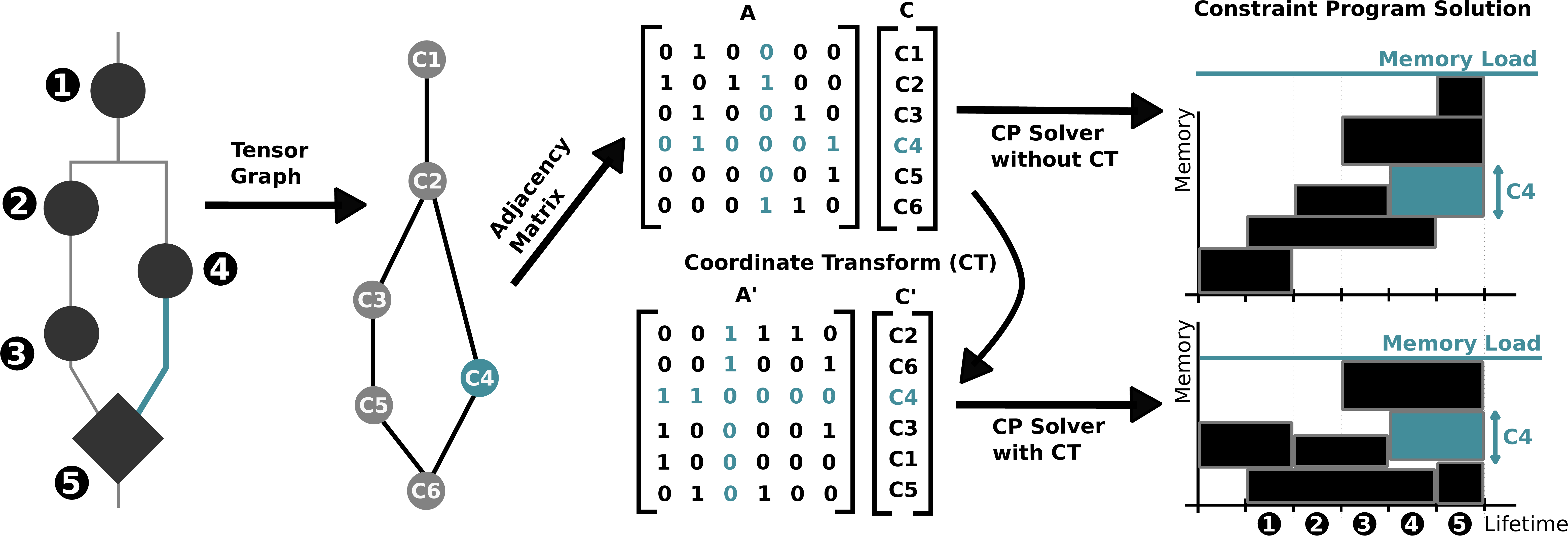}
\caption{Example of the co-optimization of tiling and static memory allocation algorithm for one memory level in Deeploy. First, the lifetime of each tensor in the graph is calculated under the execution schedule shown on the left. Next, the memory scheduler constructs an adjacency matrix of the tensor graph and extracts the cost vector from the tile constraint flow shown in the middle. Finally, Deeploy applies a coordinate transform within the \gls{cp}. On the right-hand side, the 2D bin packing solution is presented with the naive solution on top, and the solution found by Deeploy is shown below.}
\label{fig:tiling}
\end{center}
\end{figure*}

\subsection{Midend}

The second stage of Deeploy's execution flow, the \middleware{}, receives the platform-specific \gls{onnx} graph and the kernel assignment for each operator from the \frontend{}. The \middleware{}'s purpose is to perform all optimization operations required to generate low-level optimized C code for the target platform in the \backend{}. The \middleware{} is divided into two optimization steps: \memorylevelannotation{} and \tiling{}. To model the \gls{cp} used to compute the tiling and static memory allocation solution, Deeploy uses Google's ORTools. 

\subsubsection{Memory Level Annotation}

The memory level annotation step annotates every buffer in the compilation context with a memory hierarchy level. The motivation for defining the storage location of every tensor is to model code generation constraints closely to the hardware; most embedded systems designed for \tinyml{} applications use multiple memory or cache levels~\cite{prasad_siracusa_2023, ueyoshi_diana_2022} to optimize the trade-off between storage density and memory access latency. While Deeploy supports the tiling of buffers, directly assigning buffers' memory levels to lower cache levels can lead to performance improvements. When targeting accelerators that would otherwise be limited by the available bandwidth towards higher-level caches, controlling memory allocation has a significant performance impact~\cite{prasad_siracusa_2023}. 

\subsubsection{Tiling}

The second processing step in the \middleware{} is \tiling{}. For every kernel template chosen in the \frontend{}, the target platform must specify a \gls{tc}.
The \gls{tc} models the geometric and platform-specific constraints for tiling an operator. 
For a tiling solution to be correct, all geometric constraints must hold. For example, the spatial dimensions of a softmax activation's output tile must be the same as its input tile's dimensions. As such, geometric constraints do not depend on the implementation of an operator.
While it is possible to tile large tensor operators down to single instructions when targeting processor cores, the same does not hold for accelerators. Specifying \glspl{tc} and platform-specific constraints on a per-kernel basis is especially important for handling the tiling problem for loosely-coupled accelerators since they typically only support specific dimensions to be tiled, owing to their specialized datapaths~\cite{ueyoshi_diana_2022, prasad_siracusa_2023}. 

Similarly to the \typeinference{} flow, the \gls{tcf} is applied top-to-bottom through the execution schedule of the network, adding the geometric and platform-specific tile constraints of every operator to the \gls{cp}. Furthermore, the \gls{tcf} adds one symbolic variable per dimension per tensor in the network to the \gls{cp} and a symbolic variable for every tensor, representing its size as the product of all dimension variables. Using this formulation, the solution of the \gls{cp} represents the size of the largest tile. 

\subsubsection{Memory Scheduling}

After the geometrical constraints of every mapped kernel template in the network are collected and added to the \gls{cp}, Deeploy's memory scheduler calculates the lifetime of every tensor in the network over the user-provided execution schedule of the \gls{onnx} graph as shown in Figure~\ref{fig:tiling}. 
% FCONTI START
As previously mentioned, this is an essential step for autoregressive Transformers that must accommodate short-lived tensors (e.g., intermediate activations, residuals) and long-lived buffers (such as $KV$ caches).
% FCONTI END

Deeploy's memory scheduler computes a tiling path using the \platform{}'s memory hierarchy model to assign a sequence of memory transfers through the different memory levels. Using the calculated lifetimes and the tensor's size variable computed before, the memory scheduler models the problem of computing a static memory allocation schedule as a 2D bin packing problem~\cite{angiolini_efficient_2005, maas_telamalloc_2022}, where the horizontal axis represents lifetime, and the vertical axis represents memory address space.

Similar to other state-of-the-art algorithms~\cite{maas_telamalloc_2022}, Deeploy's scheduling \gls{cp} works with Tetris scheduling introduced in TetriSched~\cite{tumanov_tetrisched_2016}, where memory buffers are scheduled one after another, adding to the maximum load of each of their lifetime's bins. To solve the tiling and allocation problem in a single shot, the memory allocation of each buffer is coupled to the tiling solution, which requires expressing the order in which they are scheduled within the \gls{cp} as well. 

The first step to modeling the memory allocation problem is to pick a random schedule of memory buffers and compute the adjacency matrix $A$ of the tensor graph. We collect the memory size of each buffer, represented as an integer variable of the \gls{cp}, in a cost vector $C$. For any permutation matrix $P$, $A' = P \times A \times P^T$ is a valid adjacency matrix with associated cost vector $C' = P \times C$. A valid $N \times N$ permutation matrix can be expressed as:
\begin{align*}
p_{i,j} & \in [0,1] &\forall i,j \in [0,N-1] \\
\sum_{i=0}^{N-1} p_{i,j} & = 1 &\forall j \in [0,N-1] \\
\sum_{i=0}^{N-1} p_{j,i} & = 1 &\forall j \in [0,N-1]
\end{align*}

Next, the total memory load is computed iteratively using $A'\, \&\, C'$: since we use Tetris scheduling, we add each buffer's memory size to the size of the last scheduled buffer whose lifetime overlaps. We use a vector of intermediate variables containing one entry for each buffer, $H$, representing the memory load in the lifetime region of each buffer. The vector $H$ is computed as follows:
\begin{align*}
H_{0} & = 0 \\
H_{j} & = \text{max}_{i=0...j-1}\left( A'[j,i] \cdot H_{i}\right) + C'_{j}
\end{align*}
The total worst-case memory load for all execution steps is then computed as memory load $ = \text{max}_{i=0...N}\left( H_{i} \right)$. 

In contrast to other static memory schedule algorithms, which focus on calculating an optimal solution for memory blocks of fixed size, our algorithm combines the constraints on tile sizes and memory layout calculation into a single \gls{cp}; this allows Deeploy to simultaneously optimize static memory allocation as well as tile sizing to control memory use during the entire inference process, which is critical to matching the memory constraints of extreme-edge \glspl{soc} with the complex buffer lifetime requirements of Transformers.
An overview of the co-constrained tiling and static memory allocation algorithm is shown in Figure~\ref{fig:tiling}.

\begin{figure}
\begin{center}
\includegraphics[width=0.975\linewidth]{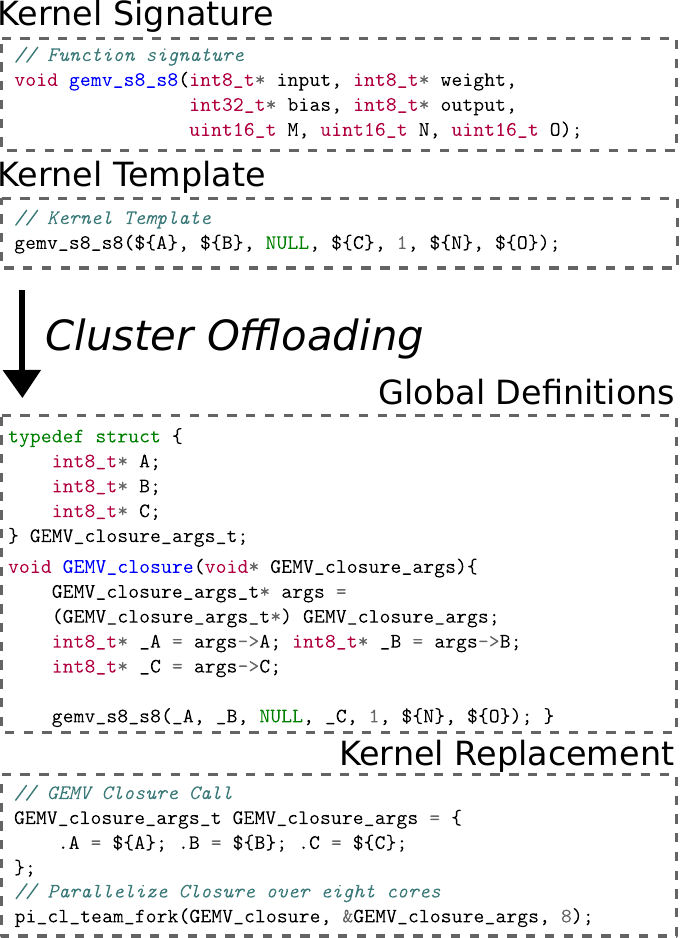}
\caption{Bottom-up offloading closure generation for a GEMV kernel. All arguments that refer to non-global \variablebuffer{}s or \constantbuffer{}s are captured and used to generate a closure struct typedef and a closure function that unpacks the argument struct and calls the original kernel. Finally, the kernel template is replaced with a function \textit{pi\_cl\_team\_fork}, which takes the newly generated closure as an argument and offloads its execution to all eight cluster cores.}
\label{fig:closure}
\end{center}
\end{figure}

\subsection{Backend}\label{sec:backend}
Every kernel template picked in the \frontend{} is assigned a list of code generation passes by the \platform{}. Each code generation pass operates on a code segment, starting from the original kernel template, and may add to or modify its code segment. Besides enabling integration of custom passes, Deeploy offers standard code generation passes required for generating correct code, e.g., memory allocation and deallocation generation, which inserts calls to heap-based allocators or sets pointers to predefined memory locations calculated during \tiling{}. 

An essential set of code generation passes is centered around generating closures for code segments. In the context of Deeploy, closure generation consists of three parts: the closure function itself, which encapsulates a code segment; the closure environment, which contains every free variable used within the code segment and must be passed to the closure function; and the closure invocation, which is either an offloading function or a call to the closure function.

\begin{figure*}
\begin{center}
\includegraphics[width=0.975\linewidth]{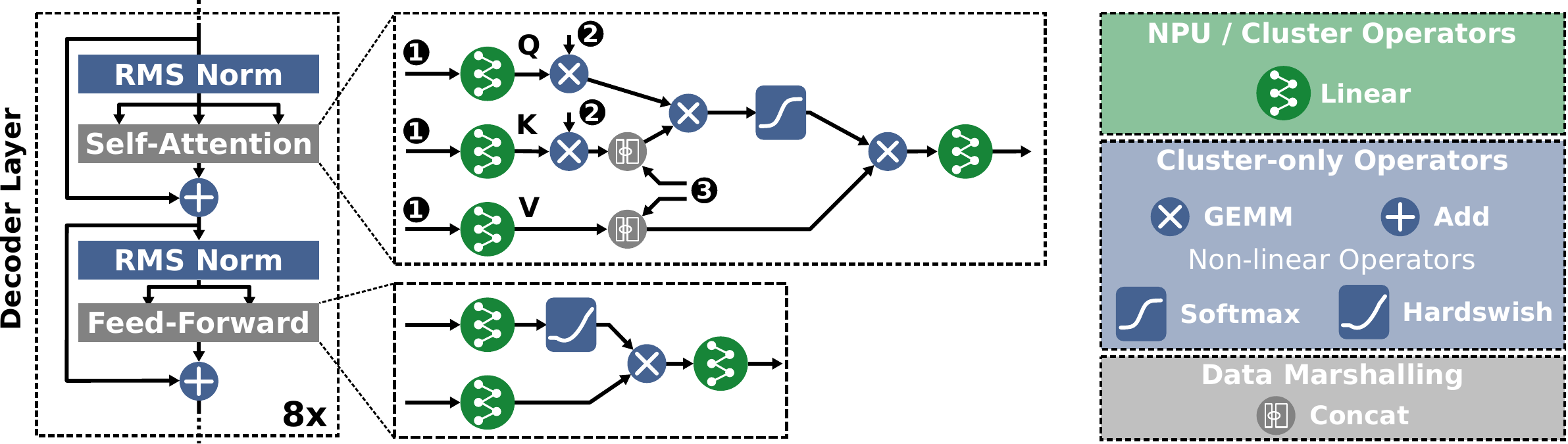}
\caption{Overview of the Llama model deployed in this work. The eight decoder layers of the model are shown on the left and consist of an \textit{RMSNorm} - \textit{Self-Attention} - \textit{RMSNorm} - \textit{Feed-Forward} layer stack. Input \circled{1} in the self-attention inset corresponds to the token input. Input \circled{2} corresponds to the rotational embedding used in Llama models. Input \circled{3} are the $KV$ cache inputs used during autoregressive inference. Notably, during autoregressive inference, the new row of the $K$ and $V$ matrices computed on the input token are appended to the $KV$ cache.}

\label{fig:tinyllama}
\end{center}
\end{figure*}

Deeploy implements closures as standard C functions by generating a function call around the target code segment and passing the closure environment as a struct pointer. Deeploy captures the relevant free variable expressions by analyzing the \gls{ast} of the underlying code segment using the Mako templating library \cite{_mako_}; since the function signature of the kernel template is known to Deeploy, it can extract arguments used in the kernel template that refer to local buffers, and pass them to the closure using an argument struct.
% Prev
%During code generation, the closure generation pass hoists the closure function definition into the global context and inserts code for constructing the argument struct, and the function call to the hoisted closure, which is returned as the new code segment for subsequent code generation passes. 
% LMACAN Rewrite
During code generation, the closure generation pass hoists the closure function definition into the global context, inserts code for constructing the argument struct and returns the function call to the hoisted closure as the new code segment for subsequent code generation passes. 

An important application for Deeploy's closures is to facilitate operator offloading, which is required for programming processor-based accelerators like compute clusters or loosely-coupled, memory-mapped accelerators like \glspl{npu}. An example of closure generation for operator offloading to the octa-core cluster is shown in Figure~\ref{fig:closure}.

Tiling code generation is implemented as a pass as well. Deeploy supports \gls{dma} engines and uses them in tiling code generation to move tiles between different memory hierarchy levels according to the tiling solution computed in \tiling{}. To hide the latency of \gls{dma} transfers, Deeploy can configure tiling for operators to use double-buffering, which constrains the tiling solution to reserve twice the required space for every input- and output tile. During code generation, Deeploy schedules data fetching and writeback to occur in parallel with kernel execution to minimize latency. 

\section{TinyStories Llama Model}\label{sec:tinyLlama}

As a concrete example of our deployment flow for next-generation \glspl{efm}, we quantize and deploy an \gls{slm} on a heterogeneous \gls{mcu}, Siracusa, introduced in Section~\ref{sec:siracusa}. We chose a Llama2 model pre-trained on the tinyStories dataset \cite{eldan_tinystories_2023} from HuggingFace\footnote{https://huggingface.co/Maykeye/TinyLLama-v0}, with a hidden size $d_{m}=64$, $h=16$ parallel attention heads, $N=8$ layers and an intermediate size $d_{ff}=256$ for the feed-forward layer. The model architecture is shown in Figure~\ref{fig:tinyllama}. Note that, however, any \gls{slm} fitting the memory constraints of the target platform can be deployed with the same flow.

Like all other decoder-based language models, the Llama model we use in this work has two fundamental inference modes, which we refer to as \autoregressivemode{} and \parallelmode{}, and generates its response in two distinct phases, the \textit{prompting phase} and \textit{generation phase}; the prompting phase ingests the initial sequence of user input tokens, whereas the generation phase generates the model's output tokens autoregressively.

\subsection{Prompting Phase}\label{sec:prompting}

Inferences follow a two-pass regime: First, the text input is translated into a sequence of tokens, typically referred to as the \prompt{}. The \prompt{} can have an arbitrary sequence length $S_{p}$, up to the size of the context window of the model.

In the first pass of the model, the \prompt{} is processed to produce the first output token. Since all tokens of the \prompt{} are available ab initio, the decoder can process them in a parallel single-shot fashion by applying \text{causal-masking} of the attention matrix~\cite{vaswani_attention_2017}. This first pass generates the first token output and the $K$ and $V$ matrices, which may be reused in the subsequent \textit{generation phase}. This process parallels the function of encoder layers used in the first Transformer models \cite{vaswani_attention_2017}.

\subsection{Generation Phase}

In the generation phase of the inference process, output tokens are generated one at a time using the previous token outputs as the model's input. While every step of the generation phase may use the same \parallelmode{} described in the previous Section, doing so would require recomputing all previous tokens' $K$ and $V$ submatrices. Therefore, the $K$ and $V$ matrices of previous inference steps are typically cached in memory to avoid the quadratic cost of recomputing them~\cite{vaswani_attention_2017}.

As the \parallelmode{} and \autoregressivemode{} require different trade-offs in memory allocation for $KV$ caching and storage of intermediate results we deploy them using separate \gls{onnx} models which reflect these trade-offs: For the \parallelmode{} we export an \gls{onnx} model with a single input and output for the token sequence and outputs for the computed $KV$ submatrices which are stored for the next generation phase.
For the \autoregressivemode{}, we use an \gls{onnx} model that additionally requires cached $KV$ submatrices. While computing outputs using $KV$ caches is significantly more efficient regarding the absolute number of operations, loading and storing the $KV$ caches induces significant data movement, and the smaller operator dimensions make the generation phase much more challenging to accelerate.

\subsection{Quantization Setup}

To quantize the \gls{slm} for deployment on extreme edge devices with integer-focused \gls{simd} processors and \gls{dnn} accelerators, we used QuantLib~\cite{spallanzani_quantlab_2022} with the \gls{tqt} algorithm for \gls{ptq}~\cite{jain_trained_2020}. 
QuantLib inserts requantization layers after operators which results in higher bitwidth outputs. Furthermore, it harmonizes scaling factors for operators like addition and concatenation and replaces various operators with their quantization-aware equivalents. Following this, we use a single token to execute \gls{ptq} over three inference epochs. Initially, we collect statistics to initialize the clipping bound for all activations and weights. At the end of the second epoch, we quantize all linear operations, and in the final epoch, we quantize non-linear operations, including Softmax and RMSNorm. 
Subsequently, the model is projected to the integer domain and exported as an \gls{onnx} graph.
To leverage the advanced hardware support for \gls{simd} operations in the PULP Cluster and Siracusa's \gls{npu}, \neureka{}, we chose to quantize all activations and weights used in matrix multiplication to \SI{8}{\bit} integer precision. We use I-BERT's approximation for Softmax~\cite{kim_ibert_2021} and the Hardswish approximation for Swish activations~\cite{howard_searching_2019}. 
Moreover, we perform all divisions in RMSNorm~\cite{zhang_root_2019} layers with \SI{32}{\bit} numerators and denominators to preserve accuracy. 

\section{Deployment Platform}\label{sec:siracusa}

This Section introduces the hardware platform used in this work as a deployment target to deploy the \gls{slm} introduced in Section~\ref{sec:tinyLlama} and goes over the \gls{npu}-specific \platform{} implementation in Deeploy.

\subsection{Siracusa}

Siracusa~\cite{prasad_siracusa_2023}, is a low-power, heterogeneous RISC-V \gls{mcu} implemented in TSMC \SI{16}{\nano\meter} technology, which is the multi-accelerator \gls{soc} targetted in this work. Siracusa is designed for efficient \gls{ai} inference, which can leverage its dedicated \gls{npu}, \neureka{}, and generalistic \gls{dsp} tasks, which can exploit both dedicated XpulpNN \gls{isa} extensions~\cite{ garofalo_xpulpnn_2020} enabling \gls{simd} processing of low-precision integers, as well as an accelerator cluster of eight RISC-V cores which enable \gls{spmd} processing.

To enable single-latency access from cluster cores to the L1 \gls{tcdm}, all cores and the 16 L1 memory banks are connected through a \gls{tcdm} interconnect using one 32-bit port each, granting a total memory bandwidth of \SI{256}{\bit \per cycle} to the compute cluster. 
The cluster's \gls{tcdm} memory banks are also accessible from the \neureka{} accelerator using 9-bank wide, \SI{288}{\bit} accesses. To manage contention on accesses to the single-ported memory banks, Siracusa integrates a lightweight, programmable access arbiter, which allows the set the maximum number of stall cycles for the accelerator; if accesses from the core-side interconnect cause accelerator access to stall for the programmed number of cycles, the arbiter will stall core accesses and grant it to \neureka{}. 

The \neureka{} accelerator uses a mixed-weight-precision bit-serial datapath, which is optimized for executing dense 3$\times$3, depthwise 3$\times$3, and dense 1$\times$1 convolution operations with \SI{8}{\bit} activations and \SIrange{2}{8}{\bit} convolution weights~\cite{prasad_siracusa_2023}. To support the bit-serial nature of the datapath, \neureka{} requires its weights to be stored in a non-standard bit-interleaved data format, which requires offline transposition, padding, and bit shuffling of \gls{cnn} weight tensors.
\neureka{} is designed as an output-stationary accelerator, opting to cache small input tiles and streaming weights.
To execute operations larger than its internal buffers, it integrates a hardware tiler with a programmable number of tiles and strides between dimensions and fixed tile sizes that match the buffer sizes.
To increase the available memory bandwidth for \neureka{}'s weights and minimize off-chip access to fetch weights, the cluster integrates a \gls{nms}, which contains two dedicated \SI{4}{\mebi\byte} memory subsystems, implemented in \gls{sram} and \gls{mram} technology respectively, which are designed to hold weights for the \neureka{} accelerator and are attached through a dedicated \SI{256}{\bit \per cycle} weight data port. 

The compute cluster and \neureka{} are located in a shared clock domain, the heterogeneous cluster, which communicates with the rest of the \gls{soc}, mainly consisting of a controller core, \SI{2}{\mebi\byte} L2 memory, and peripherals, through a \SI{64}{\bit} wide \gls{axi} bus, which can be used by a \gls{dma} integrated within the cluster, to transfer data between the L1 and L2 memories autonomously.

While Siracusa is equipped with significant computing capabilities through two dedicated accelerators and sizeable on-chip memory, deploying an advanced neural network on this device is a challenging problem. While weight storage for layers that can be executed on \neureka{} is plentiful, all other layers' activation, weight, and output tensors must be tiled to fit within \SI{256}{\kibi\byte} of L1 memory. Furthermore, memory transfers between L2 and L1 should be orchestrated using the \gls{dma} to minimize stalling.

%Operators must be tiled to map large layers commonly found in Transformers so that their inputs fit within the L1.

\begin{figure}
\begin{center}
\includegraphics[width=\linewidth]{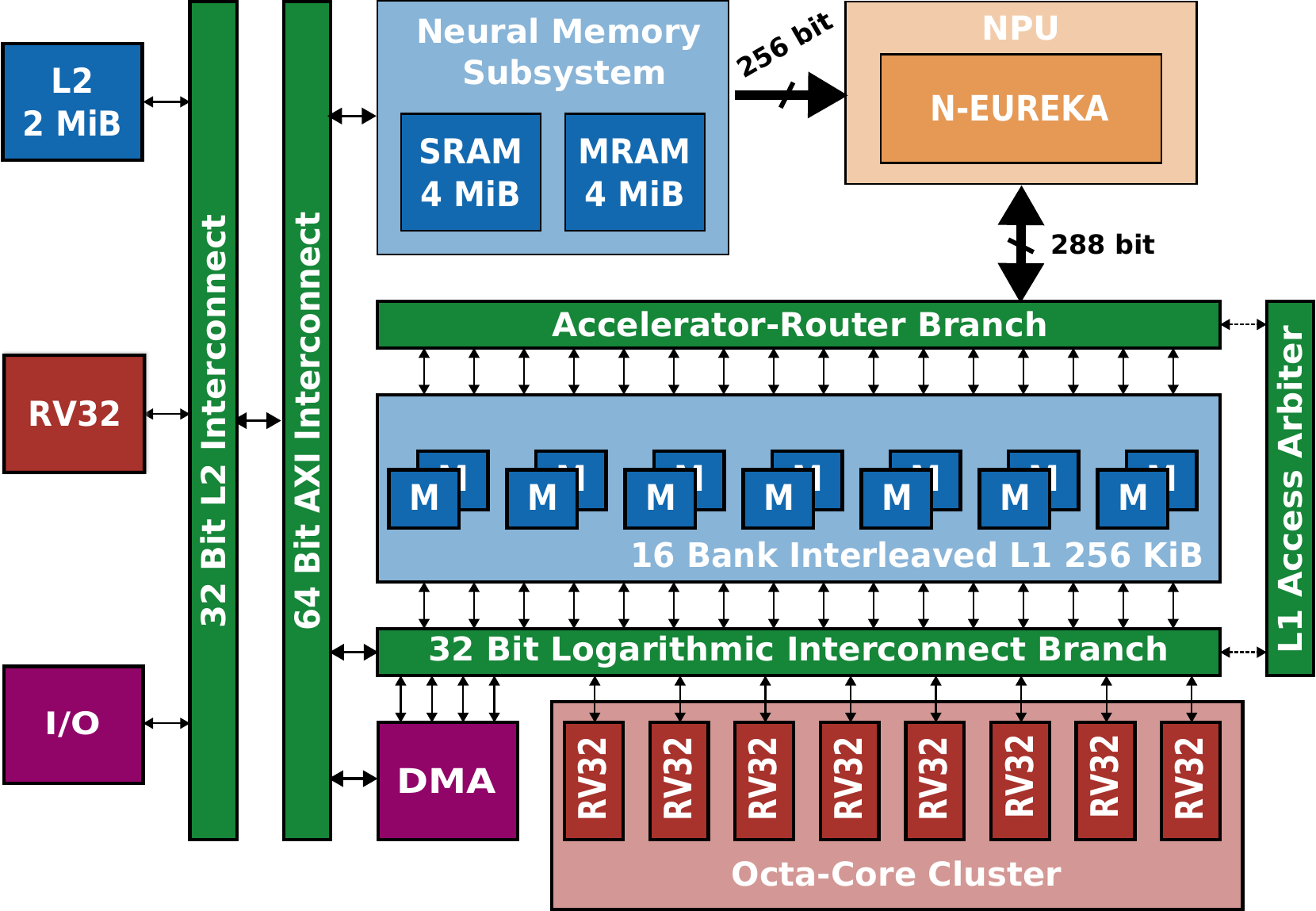}
\caption{Overview of the Siracusa \gls{soc} featuring its \gls{dsp}-enhanced octa-core RISC-V cluster and host controller (red), NPU (orange), complex memory hierarchy with two levels of scratchpad memory and a Neural Memory Subsystem (blue), two arbitrated interconnects towards the L1 memory and an \gls{axi} interconnect (green), and peripherals such as the cluster \gls{dma} and chip-level I/O (purple).}
\label{fig:heterogeneouscluster}
\end{center}
\end{figure}

\subsection{Deeploy Integration}\label{sec:pwconv}

We address the deployment challenges posed by Siracusa's heterogeneity through an augmented \platform{} model. This subsection gives an overview of the additions implemented to use Deeploy for deploying \gls{slm}s on Siracusa and, more generally, of the modifications needed to support a generic new platform in our deployment tool.

As Deeploy's core primitives are optimized kernels, we chose the PULP-NN \cite{garofalo_xpulpnn_2020} kernel library, which integrates parallel kernels as well as single-core implementations, as our target for utilizing the octa-core cluster. The PULP-NN kernels focus on efficient implementations of fused linear and quantization layers.
We support fused layers through lowering passes that match the supported operator combinations and merge them in the \frontend{} of Deeploy.
We further added fused linear operator \glspl{tc}, which add kernel-specific constraints besides providing general geometric constraints.

We implement function offloading to both the \gls{npu} and the octa-core compute cluster in Siracusa using the closure system, as detailed in Section~\ref{sec:backend}.

The \neureka{} accelerator provides greater compute capabilities than the octa-core cluster for \gls{cnn} operators, achieving a peak throughput in the range of hundreds of \si{\giga Op \per \second} for pointwise and 3$\times$3 convolutions.
Even though \glspl{slm} do not employ these types of operations, we add a custom \textit{linear layer to pointwise convolution} lowering pass that converts \gls{gemm} operators with compile-time constant weight matrices into pointwise convolutions. This method allows us to deploy all linear layers in Transformer models, as shown in Figure~\ref{fig:tinyllama}, on the \gls{npu}.

For this lowering pass, we consider \gls{gemm} operation of type $Y = \alpha AB + \beta C$, where $Q$ are appropriately integer-quantized numbers:
$$
A \in Q^{M \times N} \qquad B \in Q^{N \times O} \qquad C \in Q^{M \times O} \qquad Y \in Q^{M \times O}
$$
Similarly, we define the pointwise convolution operator as $Y = A \otimes B + C$, with the same dimension definition used in PyTorch:
\begin{align*}
A &\in Q^{H \times W \times C_{in}} \quad & B &\in Q^{C_{out} \times 1 \times 1 \times C_{in}} \\
C &\in Q^{H \times W \times C_{out}} \quad & Y &\in Q^{H \times W \times C_{out}}
\end{align*}

We map the dimensions of the pointwise convolution to those of a \gls{gemm} operation by setting $H := 1$, $W := M$, $C_{in} := N$, and $C_{out} := O$. For this mapping to succeed, the $C$ operand in the \gls{gemm} operation must be reducible to a dimension of $[\nobreak\hspace{.16667em plus .08333em}1 \times O\nobreak\hspace{.16667em plus .08333em}]$, i.e., all rows in the matrix are identical.

% Since \neureka{} uses hardware tiling with a fixed tile size, we have to preserve the minimal amount of executed \neureka{} tiles to achieve optimal performance.
% This goal impacts the tiling algorithm which we express through a dedicated TCF that constrains the tile size to be divisible by the \neureka{}'s tile size, when possible.
% Besides that, we also constraint the input channel dimension to stay untiled to fully utilize NPU's output stationary property.

Lastly, we annotate all pointwise convolution weights previously transformed from the \gls{gemm} operators to be allocated in the \gls{nms}, allowing the accelerator to leverage its significantly larger bandwidth.

% LMACAN: I am done until here

\begin{table*}
\centering
    \revA{
    \caption{Compiler performance metrics for the 128th autoregressive inference step using the \scenarioFour{} scenario with varying numbers of decoder layers}\label{tab:compiler_metrics}
    \begin{tabularx}{\linewidth}{ X r|r|r|r|r|r|r|r}
         Number of decoder layers & 1 & 2 & 3 & 4 & 5 & 6 & 7 & 8 \\ \hline
         Number of operators & 32 & 64 & 96 & 128 & 160 & 192 & 224 & 256 \\ 
         \hline
         Deeploy  compilation time [s] & 1 & 2.65 & 5.17 & 7.92 & 11.72 & 16.9 & 22.25 & 28 \\  
         \hline
         Text section size [kB] & 42.4 & 61.8 & 84.4 & 105.7 & 128.1 & 150.1 & 171.5 & 194.8 \\ 
         \hline
         Data section size [kB] (input and output buffers) & 10.4 & 18.6 & 26.8 & 35.0 & 43.2 & 51.4 & 59.5 & 67.7 \\ 
         \hline
         Data section size [kB] (requantization parameters) & 7.0 & 15.3 & 23.6 & 31.9 & 40.2 & 48.5 & 57.8 & 66.1 \\ 
         \hline
         Weight memory section size [kB] (pointwise convolution parameters) & 64 & 128 & 192 & 256 & 320 & 384 & 448 & 512 \\ 
    \end{tabularx}
    }
\end{table*}

\subsection{Deployment Setup}

As explained in Section~\ref{sec:tinyLlama}, the dual inference modes of decoder-only models require different deployment strategies, as the \autoregressivemode{} requires significant memory for $KV$ caching. We deploy two model prototypes to accommodate this difference, one for \autoregressivemode{} and one for \parallelmode{}.
%\todo{same as before, we can think to re-target a bit the Section to consider the new micro-benchmarking results}

The \autoregressivemode{} model uses additional network inputs corresponding to the previous sequences' $KV$ caches. Other than that, the deployment setup between both models is equal. We allocate all graph inputs and outputs as global \variablebuffer{}s in Siracusa's L2
memory, and annotate all local \variablebuffer{}s modeling intermediate tensors in L2 as well. In deployment scenarios that use Siracusa's \gls{nms}, we allocate all linear layer weights in the \gls{nms} but use L2 for all activations. 

Unless stated differently, all network operators are executed on the cluster and use Deeploy's \gls{tcf} to generate tiled inference code, which orchestrates transfers of input, weight, and output tensors between the L2 memory and the L1 memory. For operators executed on the \gls{npu}, weights are stored in the \gls{nms} in their entirety and ingested by the accelerator without moving them into L1 first, leveraging the increased available bandwidth from the \gls{nms}.

% \begin{figure}
% \begin{center}
% \includegraphics[width=\linewidth]{fig/gemm2pw.pdf}
% \caption{Gemm to Pointwise Convolution}
% \label{fig:gemm2pw}
% \end{center}
% \end{figure}

\begin{figure*}
\begin{minipage}{.45\textwidth}
\begin{center}
\includegraphics[width=\linewidth]{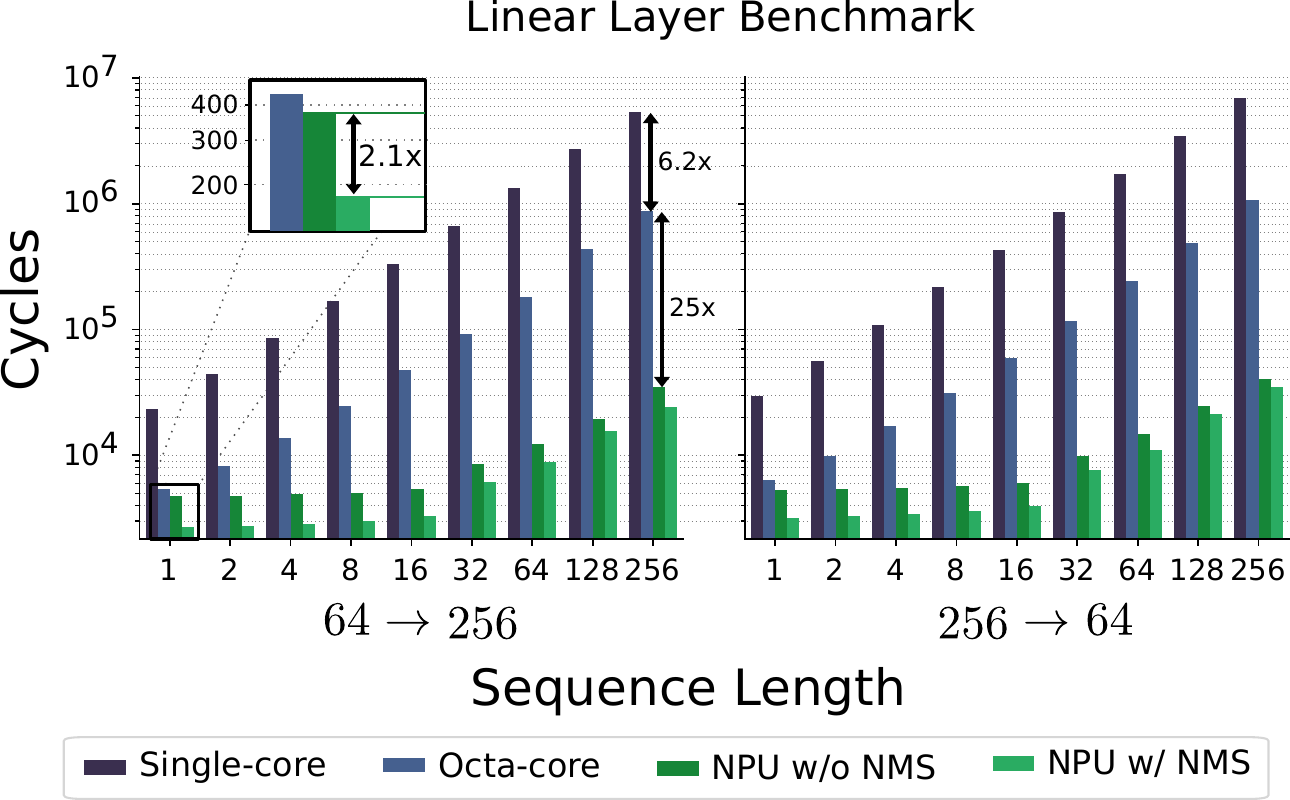}
\caption{Performance results for linear layer operators offloaded on \neureka{} using Deeploy code generation. The highlighted inset shows that the \gls{nms}' added storage and bandwidth leads to performance gains of up to 2.1\,$\times$  in memory-bound operator configurations. In large linear layer configurations, the speedup achieved by the \gls{npu} is 25\,$\times$ compared to the octa-core implementation, and another 1.6\,$\times$ when using the \gls{nms} for weights.}
\label{fig:benchmark}
\end{center}
\end{minipage}%
\hspace{.075\textwidth}%
\begin{minipage}{.45\textwidth}
\begin{center}
\includegraphics[width=\linewidth]{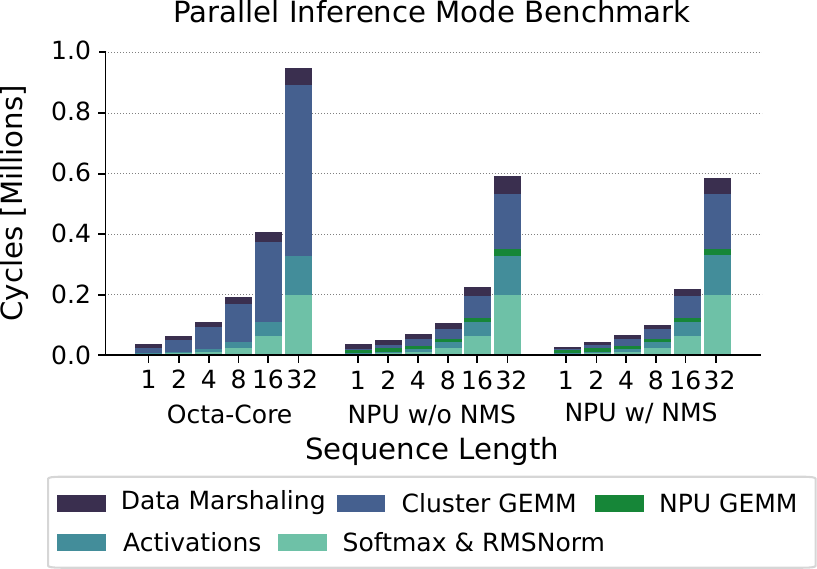}
\caption{\revA{Cycle breakdown of parallel inference in the studied \gls{slm}.} Due to the larger contribution of operations from matrix multiplications using the \gls{nms} performance of offloaded \gls{gemm} operators increases by 17.8\,$\times$, and end-end-performance improves by 61\,\% for sequence length 32 while maintaining low overheads of only 9\,\%, even when fully leveraging both the cluster and \gls{npu}.}
\label{fig:parallel}
\end{center}
\end{minipage}
\end{figure*}

\section{Results}\label{sec:results}

This section discusses the measurement results of deploying the TinyStories \gls{slm} on Siracusa and benchmarking results of general Transformer layers. First, we discuss the setup used to measure performance results on Siracusa. Finally, we present our benchmarking and end-to-end silicon measurements, as well as profiling experiments of our compiler.

\subsection{Deployment Evaluation Setup}
To evaluate the model's performance in autoregressive mode and for causally masked parallel inference, we measure each inference step individually with code generated by Deeploy. 
We start from empty $KV$ caches for causally masked parallel inference and process N input tokens simultaneously. We start from the $KV$ caches of the previous inference step for all experiments in autoregressive mode. 
%While Deeploy supports longer sequences, we follow the convention set by Eldan et al. and limit the context window to 256 tokens~\cite{eldan_tinystories_2023}. 
To calculate the average throughput and energy per token, we take the average over all 256 inference steps.

We report all power numbers measured on a Siracusa prototype board using a Keysight N6715C DC, supplying all operating voltages and measuring current. We perform all experiments under nominal conditions, i.e., \SI{0.8}{\volt} supply voltage and \SI{360}{\mega\hertz} operating frequency of the cluster domain. We measure power consumption for every inference by averaging the power consumption of the model run in a continuous loop.

We measure four distinct deployment scenarios: In the first scenario, \scenarioOne{}, we only generate code using a single RISC-V core. In the second scenario, \scenarioTwo{}, we generate code using all eight RISC-V cores of the cluster without using the \gls{npu}. In the third scenario, \scenarioThree{}, we generate code using all eight RISC-V cores and \neureka{} without offloading weights to the \gls{nms}. In the last scenario, \scenarioFour{}, we generate code using all eight RISC-V cores and \neureka{} with the \gls{nms}. We use the Siracusa \platform{} in Deeploy to generate code for all scenarios.

% \begin{figure}
% \begin{center}
% \includegraphics[width=\linewidth]{fig/tinyLlamaMicrobenchmarks.pdf}
% \caption{Performance results for \gls{gemm} operators offloaded on \neureka{} using Deeploy code generation. The highlighted inset shows that in memory-bound operator configurations, the \gls{nms}' added storage and bandwidth leads to performance gains of up to \reviewme{2.1\,$\times$}. In large \gls{gemm} configurations, the speedup achieved by the \gls{npu} is \reviewme{25\,$\times$} compared to the octa-core implementation, and another \reviewme{1.6\,$\times$} when using the \gls{nms} for weights.}
% \label{fig:benchmark}
% \end{center}
% \end{figure}

% \begin{figure}
% \begin{center}
% \includegraphics[width=\linewidth]{fig/ParallelBreakdown.pdf}
% \caption{Cycle breakdown for the end-to-end causally masked parallel inference. Due to the larger contribution of operations from matrix multiplications using the \gls{nms} performance of offloaded \gls{gemm} operators increases by \reviewme{17.5\,$\times$} and end-end-performance improves by \reviewme{77\,\%} for sequence length 16 while maintaining low overheads of only \reviewme{10\,\%}, even when fully leveraging both accelerators.}
% \label{fig:parallel}
% \end{center}
% \end{figure}

\subsection{Microbenchmarking Results}

To validate our approach of offloading \gls{gemm} operators on \neureka{}, we first measure the performance of \neureka{} and the RISC-V cluster on \gls{gemm} kernels. Specifically, we study the performance of the Q, K, and V projections in the attention layer and linear layer performance in feed-forward layers for different sequence lengths $S$ in parallel inference mode. For the Llama model we study in this paper, these projections use dimensions $256 \rightarrow 64$ and $64 \rightarrow 256$. Our measurements are shown in Figure~\ref{fig:benchmark}.

Transitioning from single-core to octa-core cluster execution, we measure a performance improvement of 6.2\,$\times$, thanks to the low-overhead parallelization on the cluster cores.
Transforming the linear layer operators into pointwise convolutions, as explained in Section~\ref{sec:pwconv}, enables execution on the \gls{npu}, which reduces latency by 25\,$\times$ compared to the octa-core implementation due to the \gls{npu}'s significant compute resources for convolution operations.
Furthermore, we reduce data movement by allocating the convolution weights to the \gls{npu}'s \gls{nms}, increasing the effective memory bandwidth available to \neureka{}. These optimizations improve performance, especially on memory-bound tasks, like linear layers in attention blocks with low sequence length, by 2.1\,$\times$ compared to \gls{npu} execution without the \gls{nms}.

We further profile the execution performance of a representative encoder layer as commonly found in non-regressive Transformer models. For our benchmarking, we chose a configuration with hidden size $d_{m}=64$ and $h=16$ parallel attention heads and an intermediate size $d_{ff}=256$, paralleling the decoder layer in Figure~\ref{fig:tinyllama}.
We measure an increase in throughput of 17.8\,$\times$ when leveraging the \gls{npu} to compute linear layers, improving end-to-end performance for encoder layers by 61\,\%. We further quantify the overheads due to tiling and data marshaling overheads, measuring an end-to-end overhead of only 9\,\%.

\subsection{Compiler Evaluation}\label{sec:compilermetrics}
\revA{To study the scalability of our code generation approach, we break down the \gls{slm} network, measuring compiler metrics for varying network depth. For each configuration, we profile code size, constant data size, and input- and output buffer size for a single inference step in \scenarioFour{}, namely the 128th autoregressive inference step. We generate all code with Deeploy and compile the resulting C Code using clang-15. Our results are shown in Table~\ref{tab:compiler_metrics}.}

\revA{We notice that while code size grows proportionally to the number of operators in the workload, the total size of the binary is dominated by weight storage in the \gls{nms}. }\revA{We further see that while compilation time grows superlinearly with the number of operators in the network, the maximum compilation time of \SI{28}{\second} does not pose a bottleneck for practical purposes. }

\begin{table}[t]
\caption{Cumulative latency and energy for a 256-step inference of the \gls{slm} on Siracusa using the \gls{npu} with \gls{nms}}
\label{tab:parallelVsAutoreg}
\begin{tabularx}{\linewidth}{Xccc}
& \multicolumn{1}{c}{\begin{tabular}[c]{@{}c@{}}Parallel\\ Inference\end{tabular}} & \multicolumn{1}{c}{\begin{tabular}[c]{@{}c@{}}Autoregressive\\ Inference\end{tabular}} & \multicolumn{1}{c}{\begin{tabular}[c]{@{}c@{}}Speedup \&\\ Energy Reduction \\ Ratio\end{tabular}} \\ \hline  
% \multicolumn{1}{l|}{Latency {[}Mcycles{]}} & 6340                                                                          & \reviewme{271}                                                                                 & \reviewme{29 $\times$ speed}                                                                                     \\ \hline
\multicolumn{1}{l}{Cumulative Latency {[}s{]}} & 17.6                                                                          & 0.75                                                                                 & 23 $\times$ faster                                                                                     \\ \hline
    \multicolumn{1}{l}{Cumulative Energy {[}mJ{]}}       &  3193                                                                         & 125                                                                                 & 26 $\times$ more efficient                                                                                    
\end{tabularx}
\end{table}

\subsection{End-to-end Deployment Results}

We thoroughly evaluate the \gls{slm} deployed on Siracusa by benchmarking the two operating phases required to execute \gls{slm}, namely the \textit{prompting phase} and the \textit{generation phase}. 

% While the memory consumption of the autoregressive regime follows a linear trend since only the data for a single token is processed, the memory consumption of the parallel regime grows quadratically with respect to the sequence length. Combined with the concurrent lifetimes of multiple tensors during inference, the parallel inference of the \gls{slm} with a sequence length of more than 32 requires more than the available L2 memory on Siracusa. For comparisons, we performed measurements on the hardware for the first 32 parallel inference steps. We extrapolated the runtime and energy for steps from 32 to 256 under the assumption of sufficient L2 memory.

Table~\ref{tab:parallelVsAutoreg} displays the cumulative runtime and energy for executing a 256-step inference in parallel mode and in autoregressive mode, where $KV$ caching is used. The autoregressive mode outperforms the parallel mode, achieving a 23\,$\times$ speedup and a 26\,$\times$ improvement in energy efficiency. These improvements directly result from avoiding the costly recomputation of $KV$ matrices. \revA{Averaging the autoregressive inference mode's cumulative latency and energy over 256 steps, we achieve an average throughput of \SI{340}{token \per \second} at an average energy cost of \SI{490}{\micro\joule \per token}.}

Since the autoregressive mode maximizes the data reuse across the whole inference process, this mode can be considered both during the \textit{prompting} and \textit{generation} phases detailed in Section~\ref{sec:tinyLlama}. However, this strategy leads to sub-optimal results as running in parallel mode for the \textit{prompting} phase enables better utilization of the \gls{npu} without excessive recomputation of $KV$ matrices, as tokens are not fed back in this phase.

\begin{figure*}
\begin{center}
\includegraphics[width=0.975\linewidth]{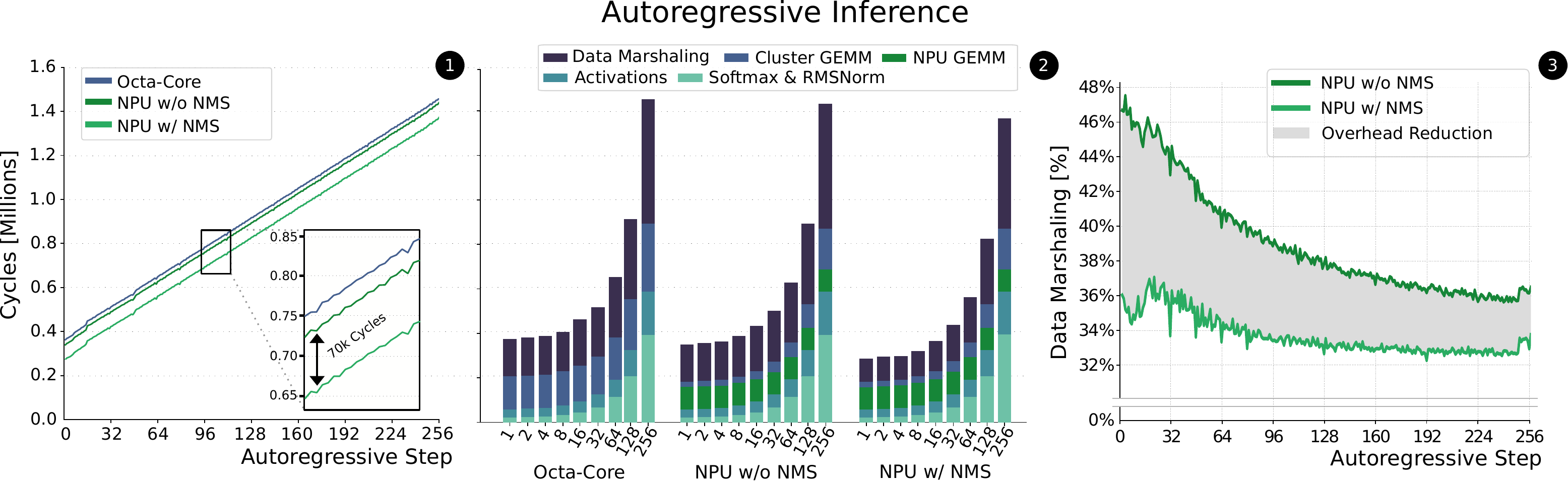}
\caption{Performance results of end-to-end autoregressive inference. Plot \circled{1} shows the runtime of each autoregressive inference step in three scenarios corresponding to \scenarioTwo{}, \scenarioThree{}, and \scenarioFour{}. The plot shows that autoregressive inference on Siracusa is highly memory-bound in all scenarios, which is due to the transfer of KV caches between L2 and L1; \scenarioFour{} reduces the runtime of every step by approximately \SI{70}{\kilo cycles} since weights are stored untiled in the \gls{nms}, reducing the required L2 - L1 data transfers. The second plot \circled{2} shows a breakdown of the runtime in the different operators of the network and data marshaling overheads. Evidently, the higher compute throughput of \neureka{} is unused due to the overall memory-boundedness. Finally, plot \circled{3} shows that the data movement overhead reduction afforded by the \gls{nms} decreases with increasing sequence lengths as the overhead of transferring KV caches increases.}
\label{fig:autorege2e}
\end{center}
\end{figure*}

\revA{The parallel inference mode's performance for the \glspl{slm} studied in this work follows the trend of the benchmark shown in Figure~\ref{fig:parallel}. While we benchmark the end-to-end performance of decoder-only models in this work, the results in Figure~\ref{fig:parallel} also apply to encoder-based transformer models, as the parallel inference mode is equivalent to encoder layer execution in such networks.}
In autoregressive mode the speedup achieved by employing the \gls{npu} is only 19\,\%, which can be attributed to the mode's smaller operator sizing, leading to stalling of the accelerator due to reconfiguration overheads. Additionally, the average proportion of time spent for data marshaling is 40\,\% for the autoregressive versus just 14\,\% for the parallel modes, underlining the memory access intensity inherent to $KV$ caching, which drastically reduces the number of computations leading to reduced arithmetic intensity.
A detailed analysis of runtime and breakdown of operator intensity for end-to-end autoregressive inference is shown in Figure~\ref{fig:autorege2e} plots \circled{1} and \circled{2}.

\subsection{Deployment Overheads}

An important metric for the quality of generated code is the utilization of the system's compute engines. To profile the quality of our code, we measured the overheads incurred by Deeploy for each autoregressive inference step in \scenarioThree{} and \scenarioFour{}, shown in Figure~\ref{fig:autorege2e}, plot \circled{3}. The main difference between the two scenarios is whether Siracusa's \gls{nms} is used for compile-time constant \gls{gemm} weights. While the reduction in overheads decreases from 33\,\% to 7\,\% with increasing sequence lengths and arithmetic intensity, the weight memory drastically reduces the relative time spent on data movement in the first steps of inference. This reduction of overheads is a crucial advantage of the bottom-up compilation approach employed by Deeploy; while other compilers might not consider low-level architectural features like memory hierarchy or only expose a simplified model, Deeploy allows complete control over memory allocation and code generation to leverage knowledge of the target architecture fully.

\subsection{Comparison with tinyML Compilers}

\revA{While we designed Deeploy to deploy state-of-the-art and emerging \glspl{slm}, we also report results on more classical \gls{cnn} and \gls{ann} workloads, as defined in the MLPerf tiny benchmark \cite{banbury_mlperf_2021}. We compare Deeploy with the state-of-the-art open-source Dory tool \cite{burrello_dory_2021} using the same open-source \gls{cnn} kernels for PULP \glspl{mcu} \cite{garofalo_xpulpnn_2020} we used in this work. To ensure a fair, compiler-focused comparison, we do not use the \gls{nms} or the \glspl{npu} of Siracusa. In this mode, both compilers only deploy cluster kernels with equivalent memory constraints. As a third data point, we add measurements of Deeploy-generated code on Siracusa when using the \gls{nms} and \gls{npu}. Our results are shown in Table~\ref{tab:tinymlperf}. We find that Deeploy generates code with an equivalent latency of Dory up to 1\,\% of variation, underlining that even though Deeploy chooses a more general compilation approach than Dory, it does not incur any performance penalties.}

\subsection{Comparison with the State-of-the-art}

Currently, most efforts on \gls{efm} deployment target models with more than a billion parameters on high-end \glspl{mpu} and embedded processors such as the I.MX95 or NVIDIA Orin or mobile phone chips, featuring multi-\si{\gibi\byte} external memories and multi-\si{\watt} power envelopes \cite{ruedas_llm_2024, chu_mobilevlm_2023}. Even though our performance and efficiency are extremely competitive, quantitative comparisons against these deployments would be unfair in our favor as we target much smaller \glspl{slm}. 

Considering \glspl{slm} in the 100s million parameters range, we compare our implementation on Siracusa with another small-scale Llama model for edge devices, MobileLLM, by Liu et al. \cite{liu_mobilellm_2024}. Liu et al. deploy a \SI{125}{\mega Parameter} \gls{slm} on an iPhone 13 featuring an A15 Bionic chip in \SI{5}{\nano\meter} technology using the highly optimized \gls{mps} backend for Apple devices, achieving a throughput of \SI{64}{Token \per \second}. While their paper does not profile the exact energy consumption of their models during inference, Liu et al. optimistically estimate the energy consumption of their setup with \SI{12.5}{\milli\joule} per token. Compared to this estimate \revA{on the iPhone 13's A15 processor, the implementation of our SLM on the Siracusa microcontroller} uses 26\,$\times$ less energy per token while achieving 5\,$\times$ more throughput, for a total 130\,$\times$ higher energy efficiency. When normalizing throughput with the number of operations per token of their network, we find that they achieve an equivalent of 4800 TinyStories Llama tokens per second. Under this estimate, our end-to-end energy efficiency on Siracusa implemented in an older \SI{16}{\nano\meter} TSMC technology node is 1.7$\,\times$ higher.

A comparison with a similar-scale (10s million parameters) model as ours is possible against the \textit{llama2.c} \cite{karpathy_llama2_2024} implementation of the TinyStories-15M model on a Samsung Galaxy Watch 4, demonstrated to achieve \SI{22.1}{Token \per \second} \cite{joeye/l[@shxf0072]_karpathy_2023} using an Exynos W920 dual-core ARM Cortex-A55 processor \cite{ifixit_samsung_2021}. Neglecting the power consumption of \gls{dram} accesses, only considering a power consumption of \SI{300}{\milli\watt} per core in Samsung \SI{5}{\nano\meter} technology \cite{frumusanu_snapdragon_2021}, we estimate the power consumption during inference as \SI{600}{\milli\watt}. Under this assumption, the Galaxy Watch 4 achieves an energy efficiency of \SI{27}{\milli\joule} per token, 55\,$\times$ lower than ours. Normalizing for operations per token, our energy efficiency is 13.4\,$\times$ greater, even though the Exynos W920 is implemented in an advanced Samsung \SI{5}{\nano\meter} technology node.

\begin{table}
    \centering
    \revA{
    \caption{Latency results of Dory and Deeploy on the MLPerf Tiny benchmark, running on Siracusa at a clock frequency of \SI{360}{\mega\hertz}.}\label{tab:tinymlperf}
    \begin{tabularx}{\linewidth}{X|c|c|c}
         Benchmark & \makecell{Siracusa w/o NPU \\ Dory} & \makecell{Siracusa w/o NPU \\ Deeploy} & \makecell{Siracusa w/ NPU \\ Deeploy}
         \\ \hline
         DS-CNN & \SI{1.4}{\milli\second} & 1.4\,\si{\milli\second} & 0.39\,\si{\milli\second} \\
         MobileNetv1 & \SI{5.6}{\milli\second} & 5.6\,\si{\milli\second} & 0.69\,\si{\milli\second} \\
         ResNet & \SI{3.7}{\milli\second} & 3.7\,\si{\milli\second} & 0.60\,\si{\milli\second} \\
         ToyAdmos & \SI{0.24}{\milli\second} & 0.24\,\si{\milli\second} & 0.11\,\si{\milli\second}
    \end{tabularx}
    }
\end{table}
\section{Conclusion}\label{sec:conclusion}

In this work, we presented Deeploy, a novel compiler for \glspl{dnn} allowing broad customizability of deployment flows. We presented the integration of Siracusa, a heterogeneous RISC-V \gls{soc} featuring an octa-core compute cluster and an \gls{npu}. We demonstrate the deployment of a \gls{slm} trained on the TinyStories dataset on Siracusa, achieving a state-of-the-art throughput of \SI{340}{Token \per\second} at an average energy cost of \SI{490}{\micro\joule} per token in autoregressive inference mode by efficiently leveraging on-chip $KV$ caching.

We further analyzed the efficiency of our generated code via microbenchmarks, achieving data marshaling overheads of only 9\,\% on Transformer encoder layers, even when fully utilizing both cluster cores and \gls{npu} collaboratively.

Lastly, we demonstrated that while data marshaling overheads are significant in the autoregressive inference mode, the energy savings compared to executing the generation phase of \gls{slm} in parallel mode outweigh this drawback, reducing the energy cost per token by 26\,$\times$ while increasing throughput by 23\,$\times$. 

In future work, we plan to leverage Deeploy's flexibility to support emerging computer architecture innovations, such as multi-accelerator \glspl{soc} integrating \gls{cim} macros. %Furthermore, we'll investigate approaches to address the data movement challenges posed by the autoregressive inference mode more effectively.

\bibliographystyle{IEEEtran}
%\bibliography{\jobname, ./references}
\bibliography{\jobname, ./Deeploy}

\end{document}